\documentclass[10pt,twocolumn,letterpaper]{article}

\usepackage{wacv}
\usepackage{times}
\usepackage{epsfig}
\usepackage{graphicx}
\usepackage{amsmath}
\usepackage{amssymb}
\usepackage{booktabs}
\usepackage{color}

\usepackage[export]{adjustbox}
\usepackage[dvipsnames]{xcolor}
\usepackage{caption}
\usepackage{subcaption}

\def\wacvPaperID{1163} 

\wacvfinalcopy 

\ifwacvfinal
\def\assignedStartPage{9876} 
\fi

\ifwacvfinal
\usepackage[breaklinks=true,bookmarks=false]{hyperref}
\else
\usepackage[pagebackref=true,breaklinks=true,colorlinks,bookmarks=false]{hyperref}
\fi
\usepackage{cleveref}

\ifwacvfinal
\else
\fi

\begin{document}

\title{Temporally stable video segmentation without video annotations}

\author{Aharon Azulay\textsuperscript{1,2}, Tavi Halperin\textsuperscript{1}, Orestis Vantzos\textsuperscript{1}, Nadav Borenstein\textsuperscript{1}, Ofir  Bibi\textsuperscript{1}\\
1. Lightricks \,\,\, 2. The Hebrew University\\
{\tt\small \{aazulay,tavi,orestis,nborenstein,ofir\}@lightricks.com}



}

\maketitle

\thispagestyle{empty} %
\pagestyle{empty} %

\begin{abstract}
    Temporally consistent dense video annotations are scarce and hard to collect. In contrast, image segmentation datasets (and pre-trained models) are ubiquitous, and easier to label for any novel task. In this paper, we introduce a method to adapt still image segmentation models to video in an unsupervised manner, by using an optical flow-based consistency measure. To ensure that the inferred segmented videos appear more stable in practice, we verify that the consistency measure is well correlated with human judgement via a user study. Training a new multi-input multi-output decoder using this measure as a loss, together with a technique for refining current image segmentation datasets and a temporal weighted-guided filter, we observe stability improvements in the generated segmented videos with minimal loss of accuracy.
\end{abstract}

\section{Introduction}

There are two main challenges in semantic video segmentation: accuracy and temporal consistency. While the accuracy challenge is essentially shared with the corresponding task for still images, temporal consistency/stability is a challenge unique to video footage. These two challenges are interconnected; for example, perfectly accurate segmentation masks are, by definition, also perfectly consistent. The opposite, however, doesn't always hold; e.g.~a sequence of empty masks is perfectly consistent, but usually far from accurate.

Another distinctly challenging issue with video segmentation is scarcity of large-scale temporally stable annotated videos, and difficulty in collecting them for a new task. In contrast, image segmentation datasets are ubiquitous, and are relatively easy to collect. For that reason, we focus our efforts on adapting image segmentation models to video as it is a more realistic scenario for real world applications.

\begin{figure}
  \centering

\includegraphics[width=.3\linewidth, frame] {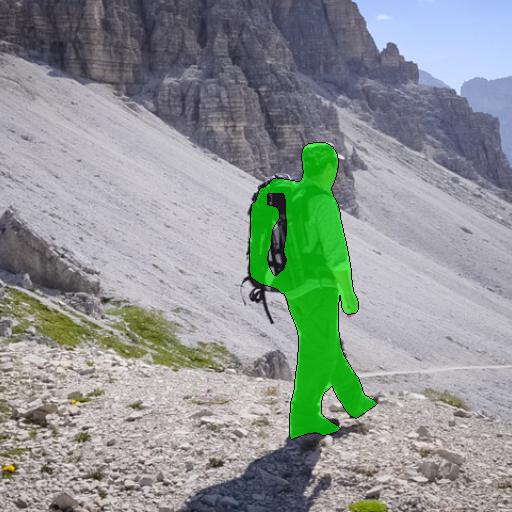} 
\includegraphics[width=.3\linewidth, frame] {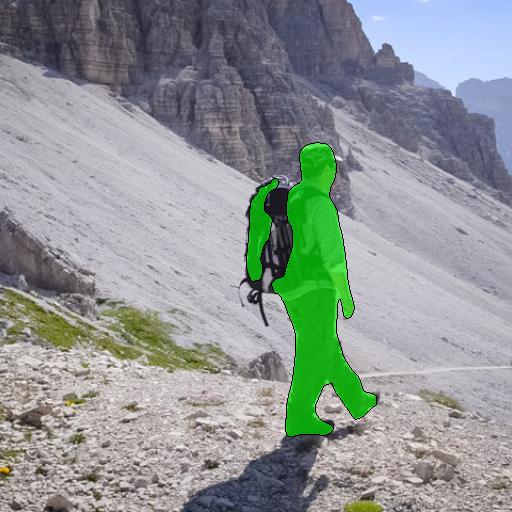}
\includegraphics[width=.3\linewidth, frame] {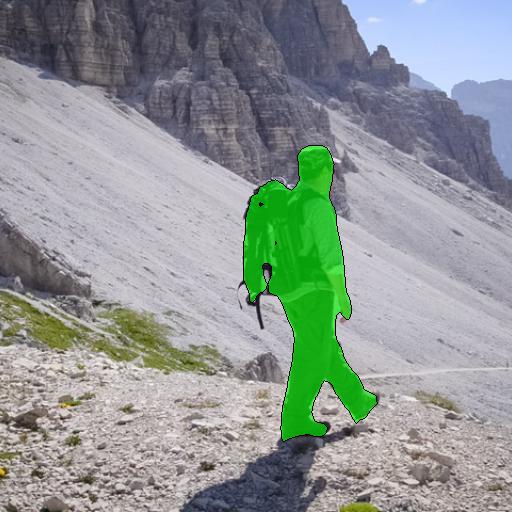}

\vspace{0.2cm}

\includegraphics[width=.3\linewidth, frame] {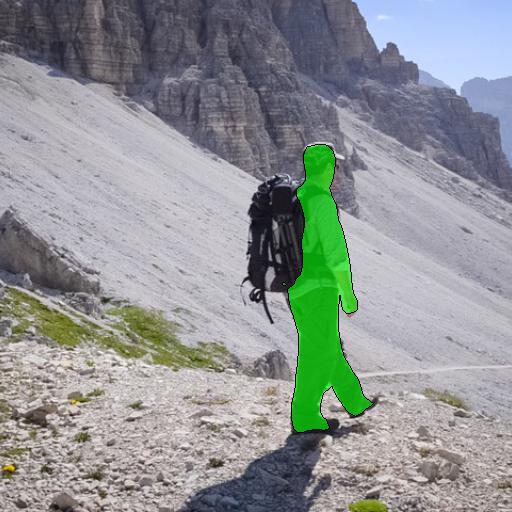} 
\includegraphics[width=.3\linewidth, frame] {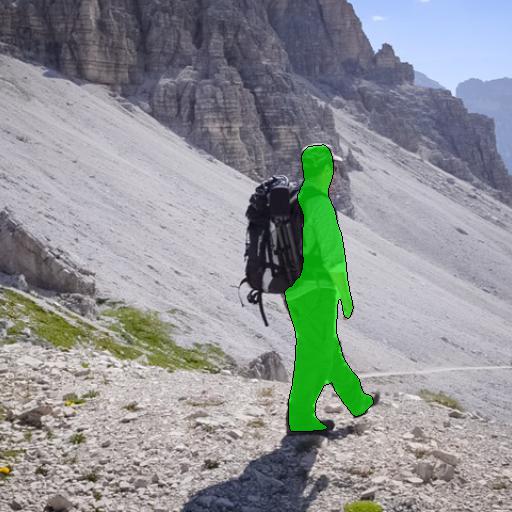}
\includegraphics[width=.3\linewidth, frame] {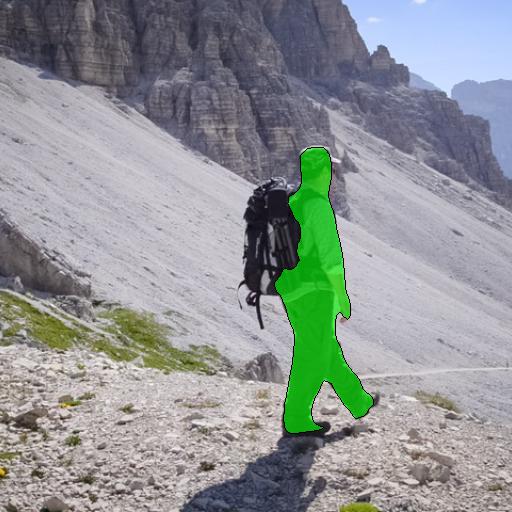}

\caption{ \textbf{Video segmentation consistency.} Top: Segmentation masks generated by running DeepLabV3 on each frame independently. Bottom: Improved segmentation consistency using our method. } 
\label{fig:teaser}
\end{figure}

Due to the essential similarity between still and video segmentation accuracy, a segmentation network trained on a large set of still images will usually generalize well to videos. But this naïve approach does not solve the consistency issue. No segmentation network is perfect, and usually the segmentation errors are only partially consistent, drawn from some error distribution. The inconsistency of the errors shows up as visible jitter of the masks. For most video-related tasks, stability is at least as important as accuracy, and users will opt to sacrifice some amount of accuracy to get visually stable results.

\begin{figure*}
  \centering

\includegraphics[width=.85\textwidth] {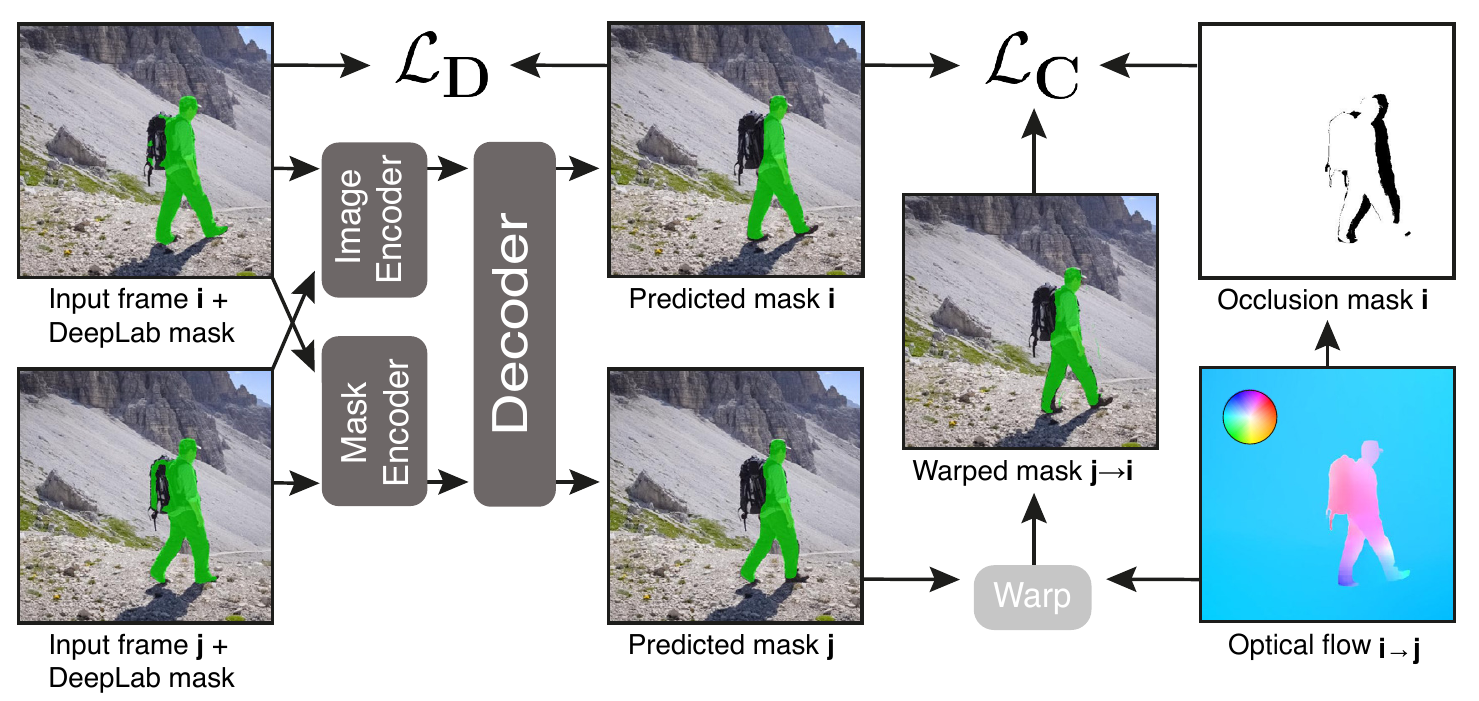}

\caption{\textbf{Multi-input pipeline overview.} Our multi-input model is fed with multiple frames (showing only two in the figure) together with the corresponding approximate masks, computed using an off-the-shelf segmentation network. It encodes the images with a shared weights encoder, and likewise for the masks, and predicts new masks via a joint decoder. We then warp each mask to every other mask in the group via the precomputed optical flow between the frames, and feed them as a supervisory signal into the consistency loss $\mathcal{L}_C$. To avoid a trivial solution, and to keep the masks semantically meaningful, we also use the data loss term $\mathcal{L}_D$ to control the amount the model can deviate from the input masks. The model output is a residual mask (not shown here for clarity), which is added to the input mask.} 
\label{fig:pipeline}
\end{figure*}

The problem of improving the temporal stability has been studied mainly in the context of video-to-video translation and video generation tasks such as style transfer~\cite{lei2020blind, lai2018learning, bonneel2015blind}. These methods are commonly referred to as 'blind', as they try to improve temporal consistency in a way that is invariant to the underlying task. Video generation tasks might be a more natural fit for these blind methods than segmentation, as there is usually no 'correct' output. For example, in video colorization, the problem is under-constrained, and there are multiple valid solutions. A temporally consistent solution is then one that does not jump between these different 'modes' (see \cite{lei2020blind} for an initial attempt to tackle the 'multiple modes' problem).

The situation in segmentation is different. There is a single correct solution (ignoring inherent ambiguity caused by motion blur, low lighting, etc.) that is also perfectly consistent. Thus, there seems to be no reason to focus specifically on consistency, which should improve anyway when accuracy improves. In practice no segmentation network is perfect, so we argue that one should strive to improve consistency regardless of accuracy. For intuition, consider a segmentation network which makes segmentation errors of only a couple of pixels around the object in each frame. If the error is consistent throughout the video, the viewer might tolerate it, while an error with similar magnitude but whose error direction is inconsistent will be more noticeable and bothersome for most users.

We propose the following contributions towards improving the temporal consistency of video semantic segmentation:

\begin{itemize}
\item We introduce an improved consistency measure  (\Cref{sec:consistency_measures}) and verify its agreement with human visual perception via an extensive user study (\Cref{sec:user_study}).
\item We propose a method for improving the temporal consistency of video segmentation without video supervision, and a method for stabilizing supervised video segmentation models, by using the consistency measure as a loss function (\Cref{sec:training}). We illustrate our pipeline in \Cref{fig:pipeline}.
\item We show further improvements to temporal consistency by refining the ground truth segmentation of still image datasets, and by post processing the network output with a temporal generalization of the weighted guided filter procedure (\Cref{sec:refined_COCO}, \Cref{sec:guided_filter}).
\end{itemize} 

\section{Related Work}
\label{sec:rel-work}

The main insight behind many approaches for improving temporal consistency in video tasks is that the motion between subsequent frames can be assumed to be small, and thus can be approximated by a simple 2D geometric transformation. The natural conclusion is that one should force the output of the model to be equivariant to small image transformations of the input. Eilertsen et al.~\cite{eilertsen2019single} put this in practice via the following loss, that forces the output of a transformed input to be the same as the transformed output of a regular input:
\begin{equation}
\mathcal{L}_{trans{\text{-}}inv} = ||f(A(x)) - A(f(x))||_2\ .
\label{eqn:stability_loss}
\end{equation}
where $A$ stands for a random affine transformation and $f$ is the image processing model. Approximating the relation between two video frames with such a simple class of geometric transformations is often not adequate, since it can not account for the variety of visual noise that appears in videos due to sensor or compression artifacts. Perhaps more importantly, affine transformations can not handle the occlusions that are inherent to the projection of 3d scenes to 2d images.

Instead of using random affine transformations in a synthetic setting, one can use a wider class of transformations based on the optical flow between real video frames. There is a series of works that take this approach for tasks like style transfer and colorization \cite{bonneel2015blind,lai2018learning, lei2020blind}, showing up in the literature under the general subject of 'blind temporal consistency'. These works mainly focus on domains where there are multiple possible solutions and temporal inconsistencies are mainly due to random switching from solution to solution between frames. For instance, colorization is a task where there are multiple distinct solutions that are equally viable (they all 'look correct') and can not be distinguished without a priori knowledge of the ground truth (the original video). In semantic segmentation, on the other hand, there is usually only one valid solution (modulo small perturbations), and even moderate deviations from it appear noticeably wrong.


Another approach is to take advantage of inter-frame continuity to share/propagate inferred information between different frames. Miksik et al.~\cite{miksik2013efficient} developed such a post-processing method based on optical flow to propagate predictions across time. Nilsson and Sminchisescu \cite{nilsson2018semantic} propagate predictions across time using a recurrent unit. Also using optical flow, Liu et al.~\cite{liu2020efficient} focused on improving temporal consistency using a more efficient, but inherently limited single frame inference. Another line of work propagates feature maps from key frames to their surroundings based on optical flow. \cite{Zhu_2017_CVPR, shelhamer2016clockwork, jain2019accel, mahasseni2017budget}. However, misalignment of key frames with nearby frames might harm accuracy relative to the original image segmentation models. Another use of optical flow is in \cite{varghese2020unsupervised}, where the authors introduce a flow-based consistency measure to evaluate, rather than directly improve, the quality of video semantic segmentation.

Going beyond optical flow, Wang et al.~\cite{wang2021temporal} proposed an attention mechanism that captures long range correlations between frames. Also using attention, Hu et al.~\cite{hu2020temporally} leveraged the temporal continuity in videos by introducing a network that distributes small sub-networks over subsequent video frames, whose features are then composed into a large feature set to be used for the segmentation.

It is also possible to propagate segmentation masks between consecutive video frames, training on still images. The key idea is to feed the network with the mask from the previous frame in order for it to infer the mask for the new frame. At training time, instead of the previous mask, the network is fed with a distorted version of the ground truth mask. A line of work \cite{perazzi2017learning, li2017video, cheng2021mivos, cheng2021stcn} showed the effect of this approach on Video Object Segmentation, a task which is similar to ours, but differs in that it requires an input mask for the first frame, to propagate to the rest of the video. Usually after pretraining on a large set of still images, they fine-tune the network on a small set of annotated videos.


\section{Consistency Measures}
\label{sec:consistency_measures}

\begin{figure}
    \centering
    \includegraphics[width=.78\columnwidth]{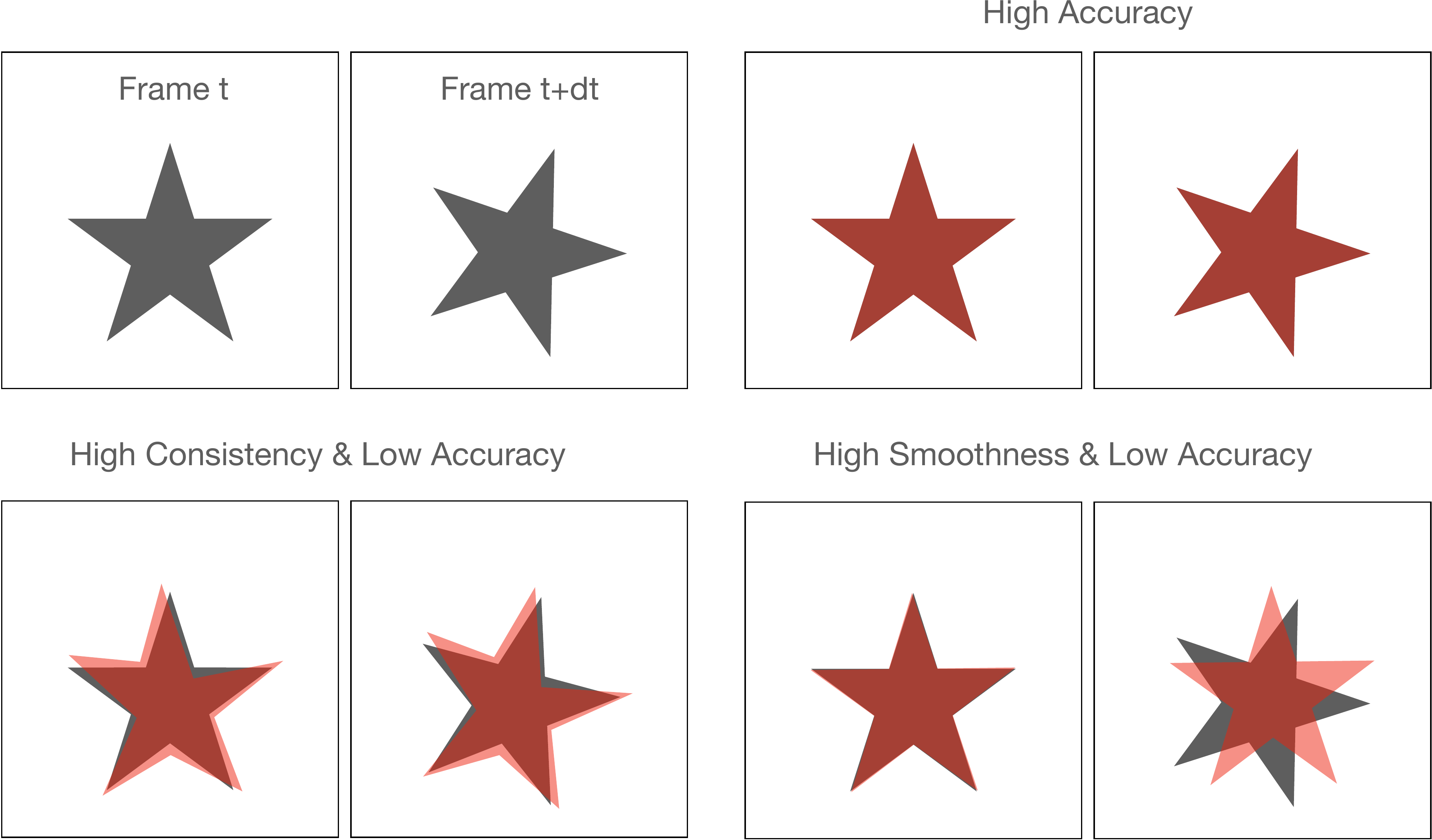}
    \caption{\textbf{Various quality measures for video segmentation.} We consider two video frames (top left), with various possible segmentations of a salient object (the star). In the best case scenario of high \emph{accuracy} (top right), the object is segmented precisely at all frames. In the case of low accuracy, we can look either at the degree of \emph{consistency}, i.e how well the mask is following the relative motion of the object (bottom left), or the degree of \emph{smoothness}, i.e how much the mask itself is changing between frames (bottom right).}
    \label{fig:measures}
\end{figure}

Most of us have an intuitive grasp of what it means for a video object segmentation to be temporally consistent. Our first step in improving consistency was to quantify that intuition. To measure the perceived segmentation consistency, we first wanted to find a metric we can calculate automatically and correlates highly with human perception, hopefully one that can be implemented as a loss in a learning based algorithm.

There are two main approaches to measure temporal consistency. The first one is based on quantifying high frequencies in the temporal domain. A typical representative of this approach is the following \emph{smoothness measure}, as presented by Eilertsen et al.~\cite{eilertsen2019single}: 
\begin{equation}
E_{smooth}(\{O_t\}_{t=1}^T) = \sum_t\lVert O_t - G_\sigma(O_t)\rVert_2 / N_{bd}
\end{equation}
%
%
%
where $O_t$ is the segmentation at time $t$, $N_{bd}$ is the number of pixels on object boundaries, and $G_\sigma$ is a Gaussian blur performed in the temporal dimension $t$. The parameter $\sigma$ is selected to be 0.15 seconds (which is roughly 5 frames for 30fps video) as described in Eilertsen et al.~\cite{eilertsen2019single}. We verified in our user study (in \Cref{sec:user_study_results}) that for YouTubeVOS videos the sigma with highest agreement with humans is about 1 frames, which corresponds to roughly 0.15 seconds, since YouTubeVOS is annotated with masks at about 6fps. This is in agreement with theory about optimal sigma~\cite{laird2006spatio}. We generalize this measure to the multi label segmentation task, by smoothing each class independently with $G_\sigma$ and merging the results with $\textrm{argmax}$. Intuitively, a segmentation mask with higher jitter will score higher (worse) on this measure, which by definition has a preference towards blurry and slowly changing masks (\Cref{fig:measures}). 

For the aforementioned reasons, we chose the second canonical approach, which uses optical flow calculation and warping, despite its computational cost, to track pixels to their target location, and verify that the segmentation remains consistent between origin and target. The quality of the optical flow is crucial for the end result of flow-based methods, hence we use GMA \cite{jiang2021learning}, a state-of-the-art deep optical flow method. On the other hand, unlike the smoothness based measures, flow based measures have the drawback of not addressing occluded pixels. Such occluded pixels are identified by a forward-backward consistency check of the optical flow \cite{sundaram2010dense, ruder2016artistic}.
Based on these optical flow and occlusion detection algorithms, we follow \cite{ruder2016artistic} in defining our \emph{optical flow-based consistency measure}. Our raw data is in the form of images (frames) $I_t$ together with the corresponding multi-class segmentations $O_t$. The segmentations are encoded in categorical form, that is each pixel is associated with a label $n\in\{0,\ldots,L\}$ where $L$ is the total number of classes. First we define a pair-wise symmetric discrepancy measure between two image-segmentation pairs:
\begin{multline}
\label{eq:pairwise_loss}
E_{pair}((I_p,O_p),(I_q,O_q)) =\\
\lVert M_{I_p\rightarrow I_q}\cdot d_{cat}(O_q, W_{I_p\rightarrow I_q}(O_p))\rVert_1 \Big/ \lVert M_{I_p\rightarrow I_q}\rVert_1+\\
\lVert M_{I_q\rightarrow I_p}\cdot d_{cat}(O_p, W_{I_q\rightarrow I_p}(O_q))\rVert_1 \Big/ \lVert M_{I_q\rightarrow I_p}\rVert_1
\end{multline}
where $W_{I\rightarrow I'}$ denotes the optical-flow derived warp between two images $I,I'$, and $M_{I\rightarrow I'}$ denotes the associated binary occlusion mask (0 for occluded pixels, 1 for non-occluded). The categorical distance $d_{cat}(\cdot,\cdot)$ takes the value 1 at pixels with the same label in both segmentations and 0 otherwise. The total measure for a video is calculated then by considering the pair-wise consistency between all pairs of frames within a temporal window of constant size: 
\begin{multline}\label{eq:Econs}
E_{cons}(\{(I_t,O_t)\}_{t=1}^{T}) =\\
\frac{1}{N}\frac{1}{\text{TK}} \sum_{|i-j|\leq K} E_{pair}((I_i,O_i),(I_j,O_j))
\end{multline}
where $N$ is a suitable normalization constant, to be discussed below. We use $K=3$ in our experiments, for a total window size $2K+1=7$.

\begin{figure}
  \centering

\includegraphics[width=.75\linewidth] {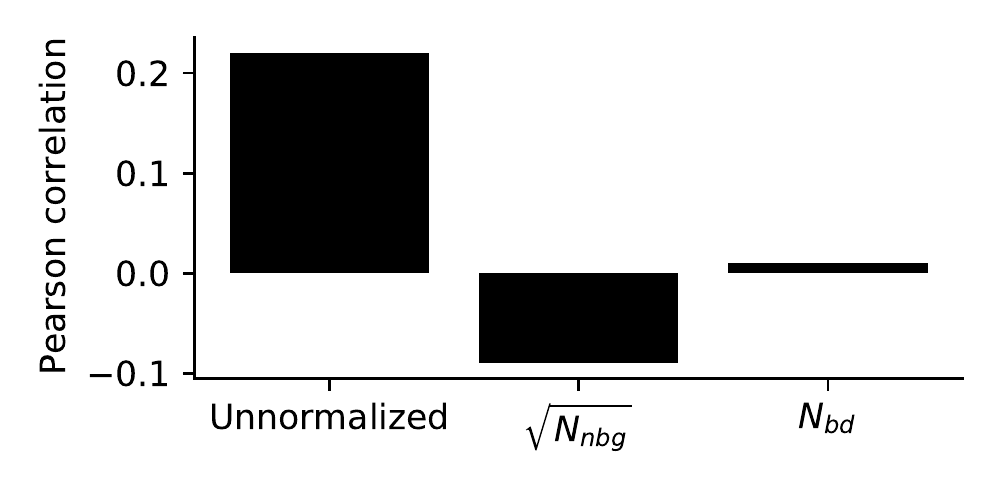} 
\vspace{-3mm}
\caption{\textbf{Consistency measure normalization}. For 1000 videos from YouTube-VOS, we found a positive correlation between the square root $\sqrt{N_{nbg}}$ of the number of non-background pixels (i.e belong to any non background class) and the consistency measure (Left). Normalizing the measure by $\sqrt{N_{nbg}}$ results in a small negative correlation (middle). When we normalize by the number of pixels on the object boundaries $N_{bd}$, this spurious correlation is suppressed.} 

\label{fig:consistency_Vs_object_size}
\end{figure}

As the measure \eqref{eq:Econs} is an average over all pixels (and not a relative measure, since we don't have a reference), it might suffer from spurious correlations with the size of the non-background objects in the masks (\Cref{fig:consistency_Vs_object_size}). The reasoning is that it is easier to be consistent for background pixels. Furthermore, we presume that the major source of inconsistencies reside in the boundaries between objects (or between an object and the background). Thus, we hypothesized that a proper normalization for the consistency measure would be by a scalar $N$ that is proportional to the object perimeter. Indeed, normalizing the consistency measure with $N=\sqrt{N_{nbg}}$, the square root of the number of non-background pixels (that is, pixels that belong to any segmentation class other than 'background'), eliminates most of the correlation with the object size (\Cref{fig:consistency_Vs_object_size}).
Because objects are often not simple geometrical shapes, their boundary might be of a fractal dimension different than 1. For that reason, instead of estimating the length of the boundaries as $N_{nbg}^{1/2}$, we counted the actual number $N_{bd}$ of pixels on the boundary of objects in each frame, and normalized with the median of $N_{bd}$ over the entire video. This further reduced the spurious correlation with object size. Furthermore, this kind of normalization slightly improved the correlation of this measure (and the smoothness measure) with human intuition (see  \Cref{sec:user_study_results}).


\begin{figure}
    \centering
    \includegraphics[width=.9\linewidth]{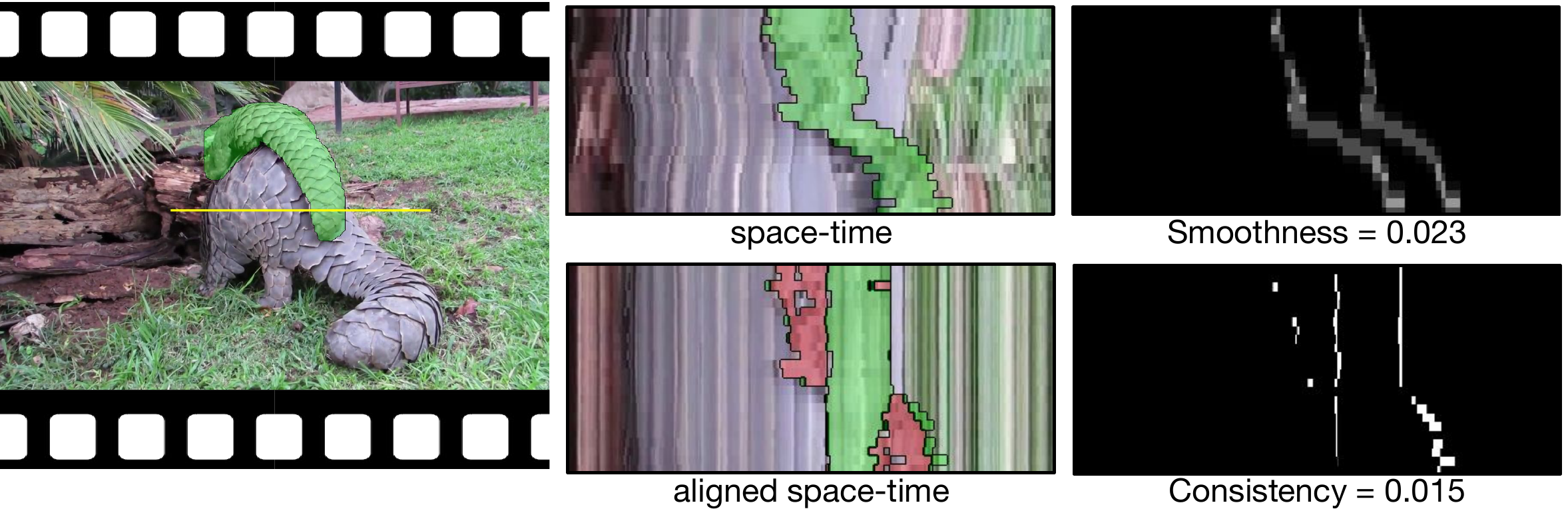}

    \includegraphics[width=.9\linewidth]{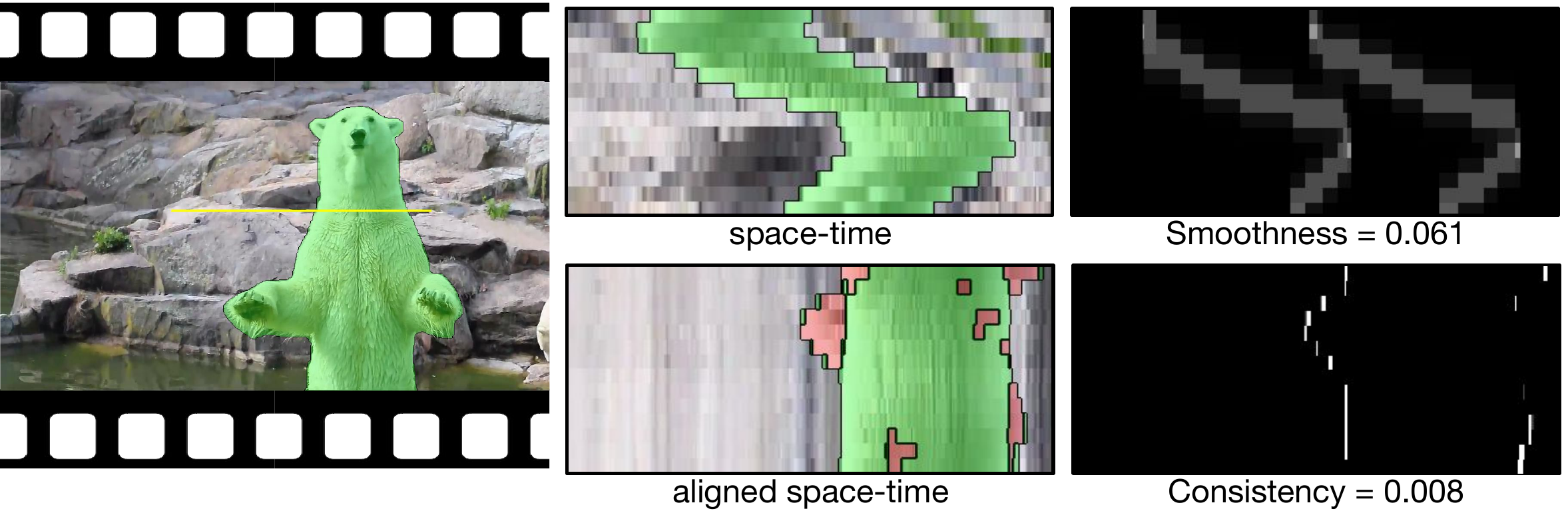}
    \vspace{-1mm}
  \caption{\textbf{Consistency vs. Smoothness.} For each example video we show a space-time slice under the yellow line, with overlaid mask. We also show the same slice where all frames were aligned based on optical flow from the middle frame of the clip. Red overlay indicates occluded pixels, computed based on forward-backward flow consistency. On the right column we show the smoothness score with the difference between smoothed and original masks, and the consistency score with the discrepancy between original and warped pixel labels. Although visually it is clear that the mask in the first example (pangolin) is less consistent, the smoothness measure prefers it over the second example (bear). Our consistency measure agrees with human judgment. Lower is better for both measurements.}
  \label{fig:consistency_vs_smoothness}
\end{figure}

To better illustrate the conceptual difference between smoothness and consistency, we show in \Cref{fig:consistency_vs_smoothness} slices in space-time from two different videos, with corresponding human-annotated segmentation masks. We then align each of the videos to one of its frames using optical flow, and compute the consistency score based on the flow. Despite some imperfections in the flow computation, the consistency score that is derived from it matches human intuition, as the human annotated mask of the pangolin is qualitatively much more jittery than the one of the bear (we include both videos in the supplementary material). The smoothness measure, however, shows inverse relation in the case of these two videos, since the bear moves a significant distance while having a precise annotation, whereas the pangolin shows smaller motion, but its annotation is more jittery.

\section{User Study}
\label{sec:user_study}
\subsection{Experimental setup}

We designed a user study to verify that our proposed consistency measure is well aligned with the human perception of consistency, by asking crowd workers to rank the consistency of the segmentation masks of $96$ videos on a Likert scale of 1-5. We paid the workers \$0.24 per labeling job and collected nine rankings for each video.

In setting up the study, we noticed that some participants tend to mix-up the concepts of consistency and accuracy. That is, they ranked the \emph{accuracy} of the segmentation masks instead of their \emph{consistency}. To mitigate the problem, we introduced ``qualification videos'' into the experiment. To construct a qualification video, we select a clip with an accurate ground-truth segmentation mask, and corrupt the mask along two separate axes -- accuracy and consistency. To corrupt the accuracy we erode $n$ pixels from the mask in a random direction, where $n$ is in $\{0, 10, 20\}$. To corrupt the consistency of the mask we use a randomly chosen local piecewise-affine transformation, where the displacement magnitude for each pixel is randomly drawn from $\{0, 4, 8\}$ and the direction is randomly chosen. Visually judging the accuracy and consistency for the corrupted set of videos, we verified that the transformations had the intended effect as we defined them. After corrupting the original video along the two axes we get a $3\times3$ grid of corrupted clips, where each clip in the grid has an accuracy rank and a consistency rank, each on a scale of 1-3. Overall, we introduced $4 \cdot 9=36$ qualification videos to the user study. Our expectation is that the ranking of crowd workers that correctly distinguish between consistency and accuracy (i.e., understand the task) should have high correlation with the \emph{consistency} value of the qualification videos, but low correlation with the \emph{accuracy} value of the videos. In contrast, the ranking of crowd workers that confuse the concepts of consistency and accuracy should have high correlation with accuracy value of the qualification videos. 

Filtering out the responses of crowd workers whose ranking of the qualification videos had higher correlation with the accuracy value of the videos than with the consistency value of the videos, the average ranking was $2.65$ and the average standard deviation for each video was $1.08$. The agreement between the qualifying workers was acceptable for such a subjective task, with mean pairwise Spearman correlation of $0.38$. We aggregated the nine 1-5 labels for each video by taking their mean.


\subsection{Results}
\label{sec:user_study_results}

\begin{figure}
\centering
  \begin{subfigure}[t]{0.19\textwidth}
    \includegraphics[width=\textwidth]{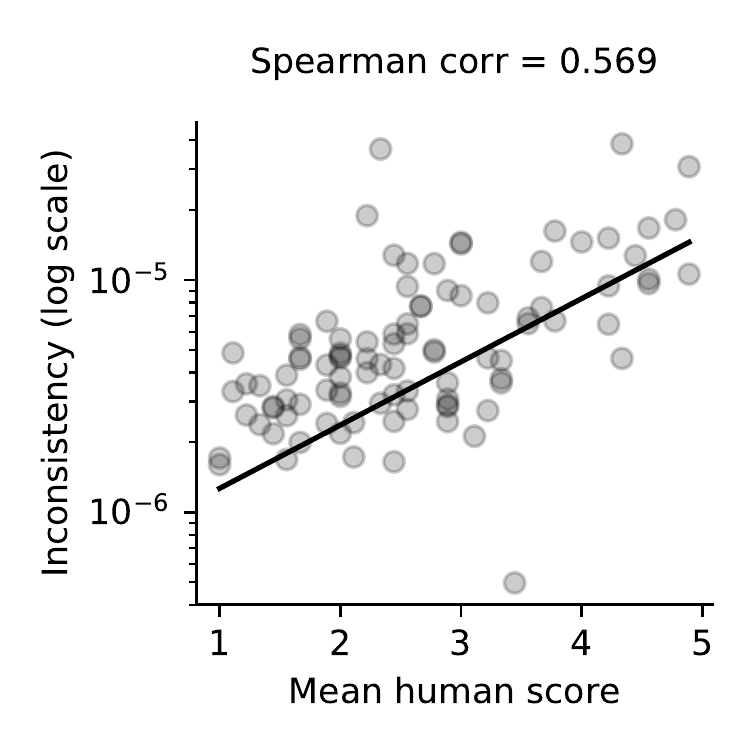}
    \label{fig:user_study_inconsistency_all}
  \end{subfigure}
  \quad
  \begin{subfigure}[t]{0.19\textwidth}
    \includegraphics[width=\textwidth]{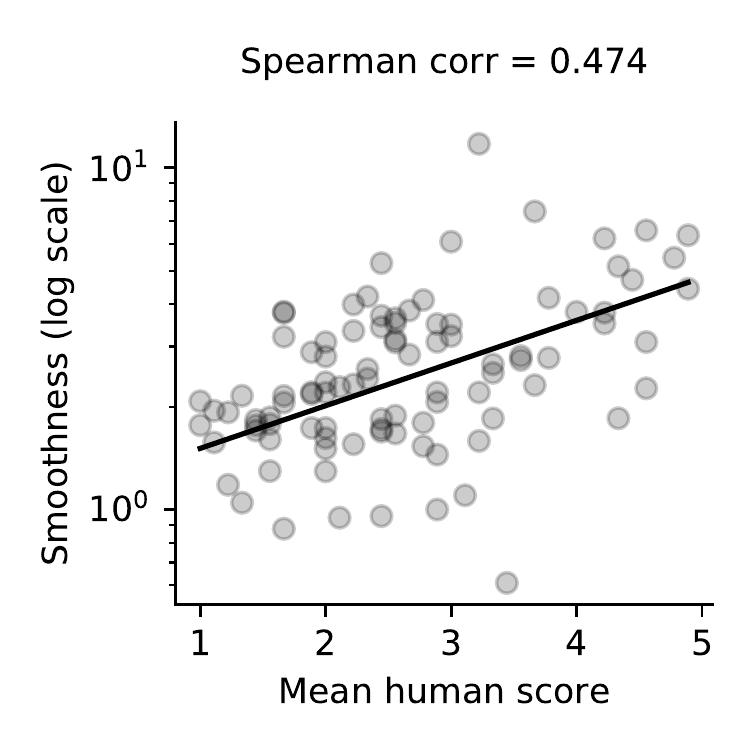}
    \label{fig:user_study_smoothness_all}
  \end{subfigure}
    \quad
  \begin{subfigure}[t]{0.19\textwidth}
    \includegraphics[width=\textwidth]{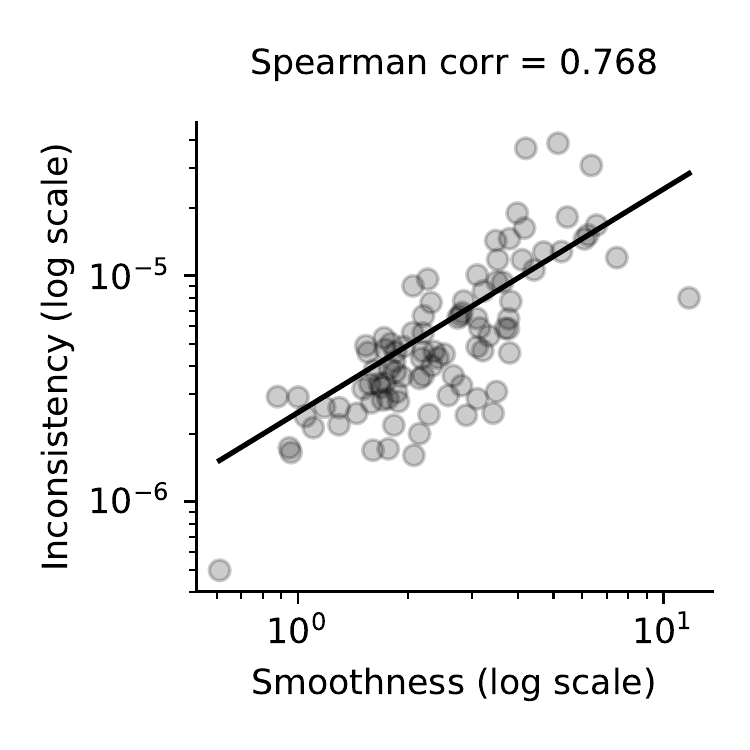}
    \label{fig:user_study_smoothness_vs_consistency}
  \end{subfigure}
      \quad
  \begin{subfigure}[t]{0.19\textwidth}
    \includegraphics[width=\textwidth]{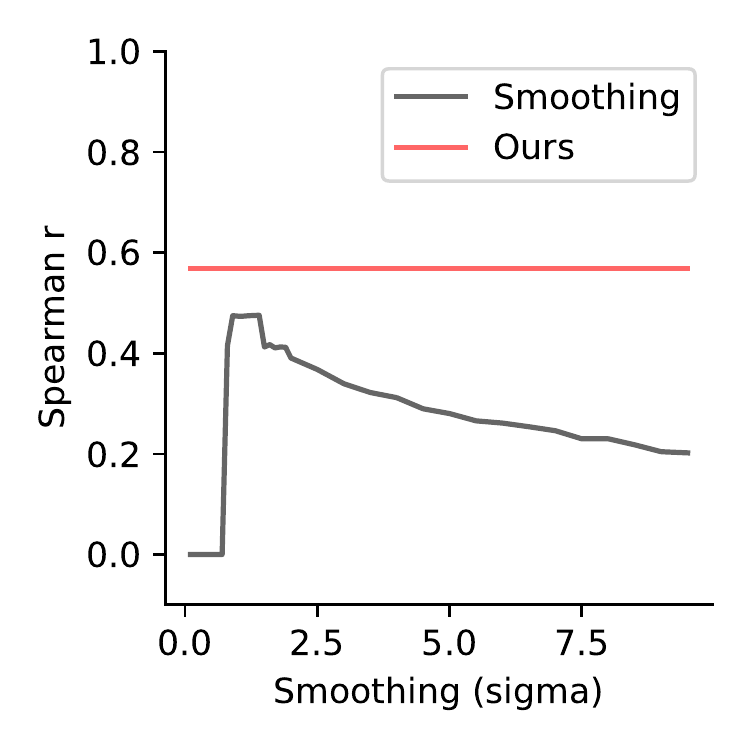}
    \label{fig:user_study_smoothness_all_sigmas}
  \end{subfigure}
  \vspace{-3mm}
  \caption{\textbf{User study.} The smoothness and consistency measures are not perfectly correlated (bottom left). The top subfigures show that our consistency measure (left) is better correlated with human judgment of consistency relative to the smoothness measure (right) \cite{eilertsen2019single}. The standard deviation for the smoothness measure was manually chosen to maximize the correlation with humans (bottom right). Videos where taken from the YouTubeVOS dataset.}
  \label{fig:user_study_inconsistency}
\end{figure}

As can be seen in \Cref{fig:user_study_inconsistency}, the proposed consistency measurement is relatively well aligned with the human rankings of consistency, reaching a Spearman correlation of $0.569$ (p-value=$1.5\mathrm{e}{-9}$). 

However, our user study demonstrates that the smoothness measure captures the human notion of consistency to a lesser extent. Indeed, as can be seen in \Cref{fig:user_study_inconsistency}, this measurement is less correlated with the rankings of crowd workers, having a lower Spearman correlation of $0.474$ (p-value=$1.08\mathrm{e}{-6}$)).


We were also interested in verifying that the normalization method described in \Cref{sec:consistency_measures} indeed improves the correlation of the inconsistency measure with human intuition of consistency. As expected, The correlation between the non-normalized version of the inconsistency measure and the human rankings was a bit lower (Spearman corr=$0.535$, p-value=$1.9\mathrm{e}{-8}$).



Given the subjective nature of the task, it is not unexpected that the participants might fail to reach consensus in their rankings for some of the videos. We inspected a random sample of such low-agreement videos (standard deviation of the nine rankings higher than $1.0$), to better understand their effect on the results of the study. We noticed that in many cases the masks in those videos were relatively consistent over the majority of the segmented object, but very inconsistent in a specific region of the object or in specific frames. Therefore, a possible explanation for the low agreement in those videos is that some participants tend to focus on the occasional artifacts, while others tend to rate the overall consistency; a difference of opinion in what constitutes a 'consistent' video. Removing the low-agreement videos from the sample, the correlation between our consistency measure and human rankings increases to $0.724$ (p-value=$1.3\mathrm{e}{-6}$). This trend applies also to the smoothness measure -- removing low-agreement videos improves its correlation score to $0.671$ (p-value=$1.39\mathrm{e}{-5}$).
\section{Training}
\label{sec:training}
\subsection{Multi-input Decoder}
\label{sec:decoder_training}

Intuitively, the temporal consistency of video segmentation can benefit with the access to information from multiple frames. However, it is not clear how to incorporate such information. Towards this end, we propose a modification of the popular DeepLabV3 \cite{chen2017rethinking} decoder to process multiple frames by increasing the number of input and output channels (\Cref{fig:pipeline}). We decided to do this modification only to the decoder for computational efficiency.


We use a standard DeepLabV3 encoder with a ResNet50 backbone to encode the frames and the segmentation outputs from an off-the-shelf DeepLabV3 semantic segmentation model. On top of these encoded representations, we train a multi-input multi-output decoder to predict the delta from the prediction of the initial DeepLabV3 model, which effectively serves as the teacher. During training, we pass randomly chosen sets of consecutive input frames to our pipeline, and we minimise the mean of the pair-wise consistency loss \eqref{eq:pairwise_loss} over every pair of input frames, together with a cross entropy data loss based on the output from the DeepLabV3 teacher. We use the Adam optimizer and a learning rate of $10^{-5}$. 

\begin{figure}
  \centering

\includegraphics[width=.30\linewidth] {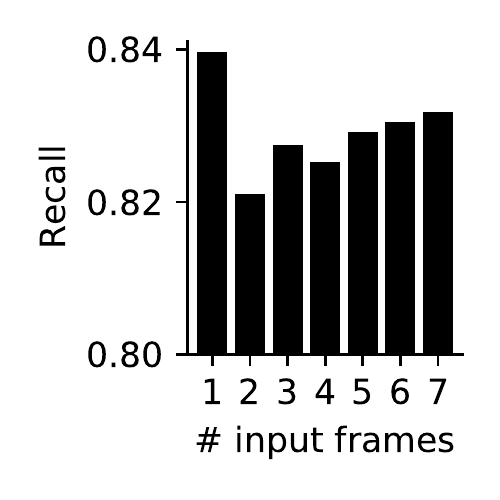}
\includegraphics[width=.30\linewidth] {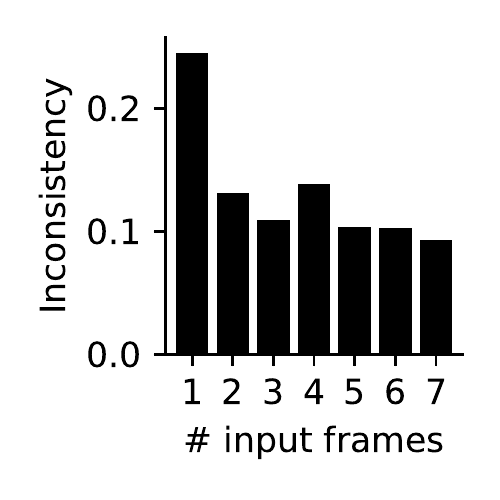}
\includegraphics[width=.30\linewidth] {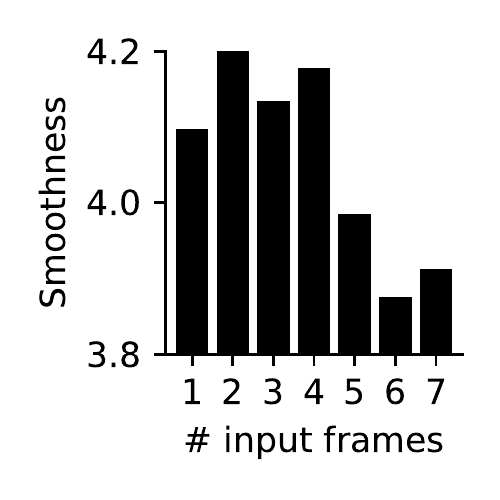}
\caption{\textbf{Effect of number of input frames.} Increasing the number of input frames for our multi-input decoder improves recall (higher is better), consistency and smoothness (lower is better). The left bar is unmodified DeepLabV3. Our 7-inputs decoder shows favourable consistency and smoothness, with only a small penalty in recall. Results are averaged over the DAVIS dataset.} 
\label{fig:ConsistencyVsN_inputs}
\end{figure}

A larger number of input frames has the advantage of providing a larger set of non-occluded pixels, as an occluded pixel in one frame might be dis-occluded in another. Moreover, it can provide multiple votes for every non-occluded pixel, correcting for some warping errors. However, increasing the number of inputs is computationally expensive, as it increases the number of optical flow computations during training quadratically, and also makes inference slower. For our experiments we used 7 input frames; see \Cref{fig:ConsistencyVsN_inputs} for the effect of different number of input frames.


\subsection{Refined Dataset}
\label{sec:refined_COCO}

\begin{figure}
  \centering

\includegraphics[width=.40\linewidth, frame] {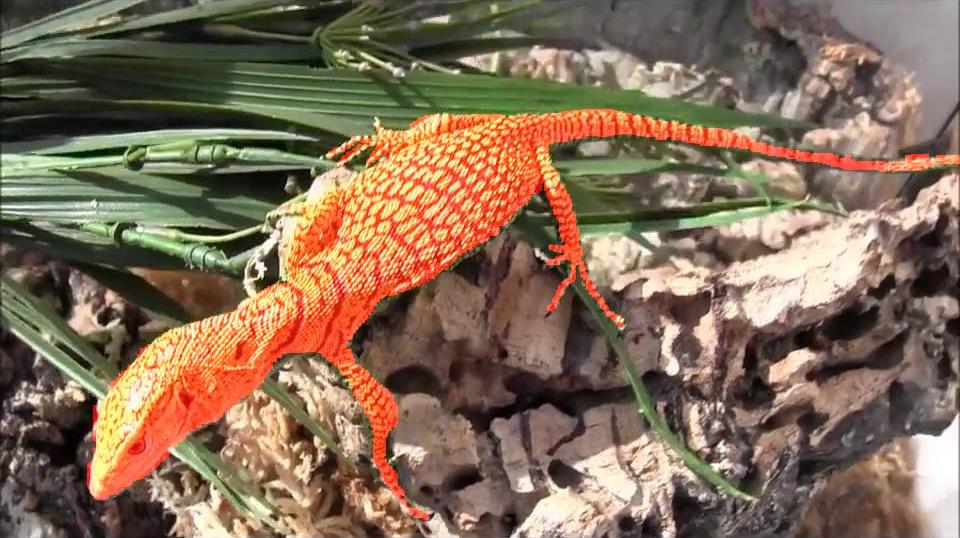} 
\includegraphics[width=.40\linewidth, frame] {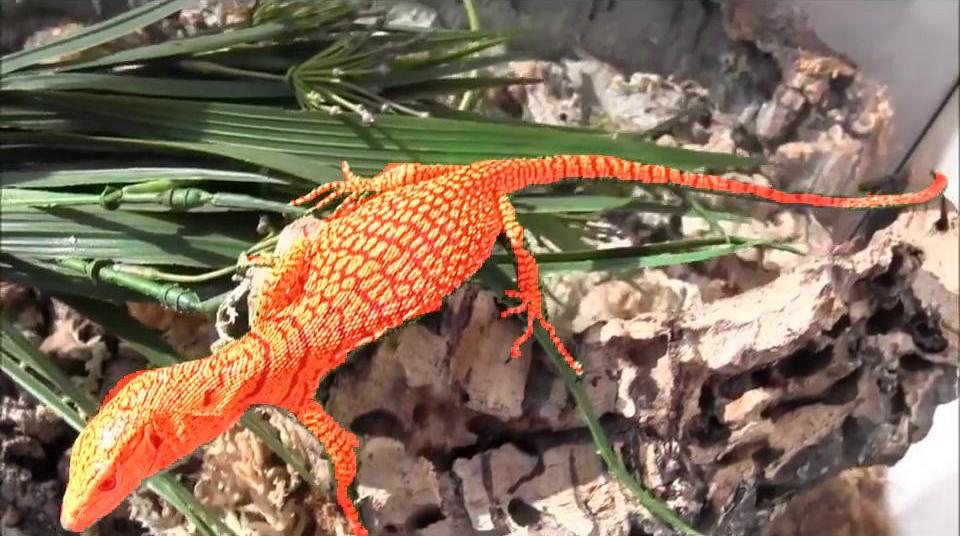}

\vspace{0.2cm}

\includegraphics[width=.40\linewidth, frame] {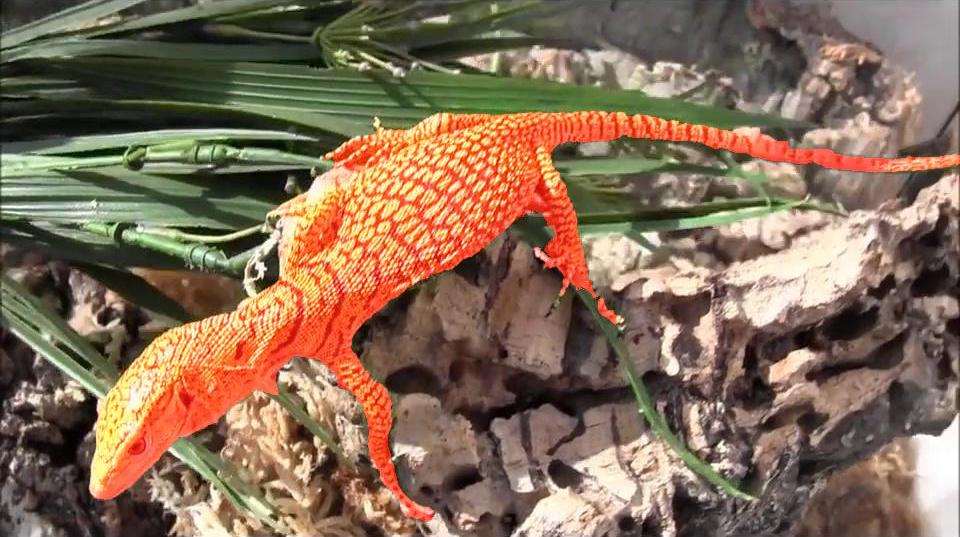} 
\includegraphics[width=.40\linewidth, frame] {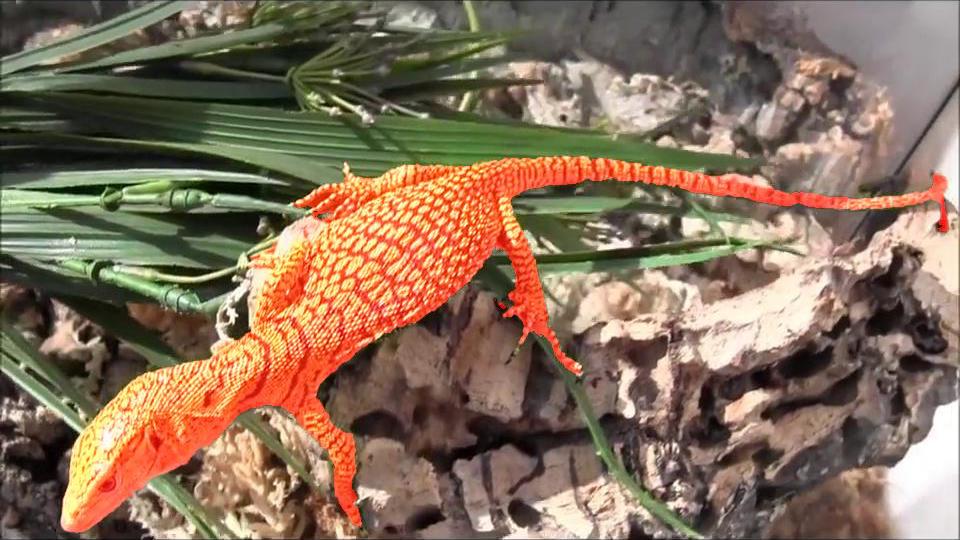}

\includegraphics[width=.43\linewidth] {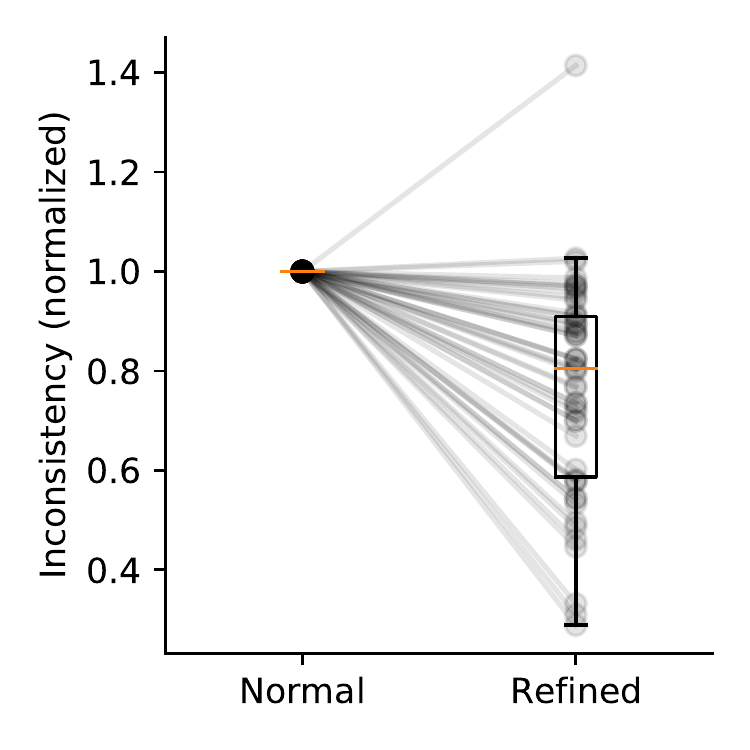} 
\includegraphics[width=.43\linewidth] {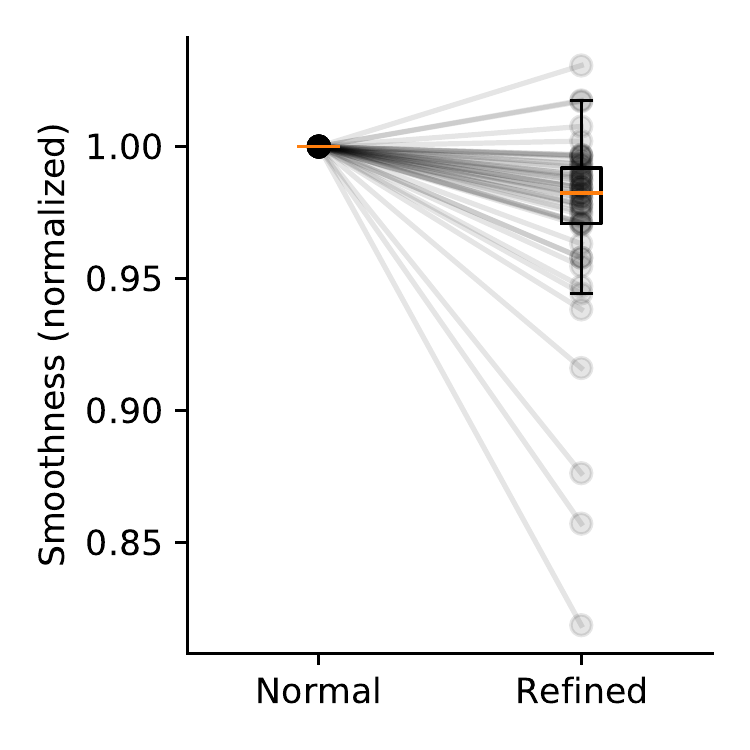}

\caption{\textbf{Refining masks improves consistency.} Top: a typical example from YouTube-VOS with small labelling inconsistencies (note around lizard head). Middle: A refined version where the mask more closely follows object edges. Bottom: both measures of temporal consistency are improved when refining the annotations (lower is better). Each point represents a video.} 
\label{fig:YoutubeVOS_refinement_consistency}
\end{figure}

The quality of the teacher affects the quality of our multi-input decoder, regardless of temporal consistency. To improve the teacher quality, we tested whether training it on a refined version of MS-COCO \cite{lin2014microsoft} will result in better segmentation masks and a better teacher. We used the latest Cascade-PSP \cite{cheng2020cascadepsp} to create a refined version of MS-COCO, to get mask edges that are more closely aligned to object boundaries (\Cref{fig:YoutubeVOS_refinement_consistency}). Note that such refinement procedures are too computationally intensive for use during inference.

We observed that training DeepLabV3 on such a refined version of MS-COCO improves temporal consistency on the DAVIS dataset, and slightly improves IoU score on a subset of 500 relabelled examples from PascalVOC  \cite{cheng2020cascadepsp}  (\Cref{tab:main}). However, training on a refined dataset slightly decreases the recall on DAVIS. We believe that this is caused by the fact that the refinement method we used (Cascade-PSP) has a tendency to drop small, background objects. Regardless, this result means that the network is prone to learning the labelling errors in the training dataset, and therefore improving labelling quality can have a substantial effect on temporal consistency when shifting the domain to video segmentation (\Cref{fig:YoutubeVOS_refinement_consistency}).

\subsection{Benchmarks}
\label{sec:benchmarks}

In \Cref{tab:main}, we show how well the proposed multi-input decoder ('Ours'), trained with our consistency loss (with 7 input frames) on DeepLabV3 representations of DAVIS video frames, performs against a number of alternatives in terms of accuracy and temporal stability. The baseline approach ('Naive') is the equivalent single-input DeepLabV3 model trained on MS-COCO, applied to the test videos frame-by-frame. We also tested an implementation of the transform-invariant technique ('TI') \cite{eilertsen2019single} discussed in \Cref{sec:consistency_measures}. We observe that the accuracy ('Recall') of our approach is comparable to the baseline, whereas the transform-invariant approach scores lower. In terms of the smoothness measure, our model scores slightly better than the baseline, whereas the TI model performs again worse. Finally for the consistency measure, the TI model improves slightly on the baseline, but our model scores significantly better.

We also tested for the effects of training on the refined COCO dataset described in \Cref{sec:refined_COCO}. Training the baseline model on the refined dataset ('Refined') led to lower recall, although the IoU score improved slightly, with small gains in smoothness and consistency. In terms of the temporal stability, training the multi-input model on the refined dataset ('Ours+Refined') led to further gains in consistency at the expense of lower recall. We believe that this is due to the tendency of the refinement network \cite{cheng2020cascadepsp} to undersegment, especially for small, background objects. This may be desirable if one is interested in segmenting the main object in the video and values stability over accuracy.

\begin{table}
  \centering
   \resizebox{1.\linewidth}{!}{

\begin{tabular}{lcccc}
    \toprule
     & Recall $\uparrow$ & IoU  $\uparrow$ & Smoothness $\downarrow$ &  Inconsistency $\downarrow$  \\
    \midrule
    Naive  & \textbf{0.839} & 0.635 & 4.09 & 0.245  \\
    TI \cite{eilertsen2019single}  & 0.760 & 0.634 & 4.40 & 0.215 \\
    Refined & 0.778 & 0.640 & \underline{4.02} & 0.223 \\
    Ours & \underline{0.831} & - & \textbf{3.91} & \underline{0.093} \\
    Ours+Refined & 0.720 & - & 4.27 & \textbf{0.075} \\
    \bottomrule
\end{tabular}
}
\vspace{-1mm}
  \caption{
      \textbf{Improving the stability of image segmentation models.} Recall was computed on the DAVIS dataset, and IoU was computed on the relabeled PascalVOC dataset \cite{cheng2020cascadepsp}. Naive is a DeepLabV3 with ResNet50 backbone trained on MS-COCO. TI stands for the Transform Invariance regularization method \cite{eilertsen2019single}. Refined is the Naive model, fine-tuned on a refined version of MS-COCO. Our method is a multi-input decoder with 7 inputs starting from the Naive model (Ours) and from the Refined model (Ours+Refined). (lower smoothness/inconsistency is better)
      }
  \label{tab:main}
\end{table}



\begin{figure}
  \centering

\includegraphics[width=.65\linewidth] {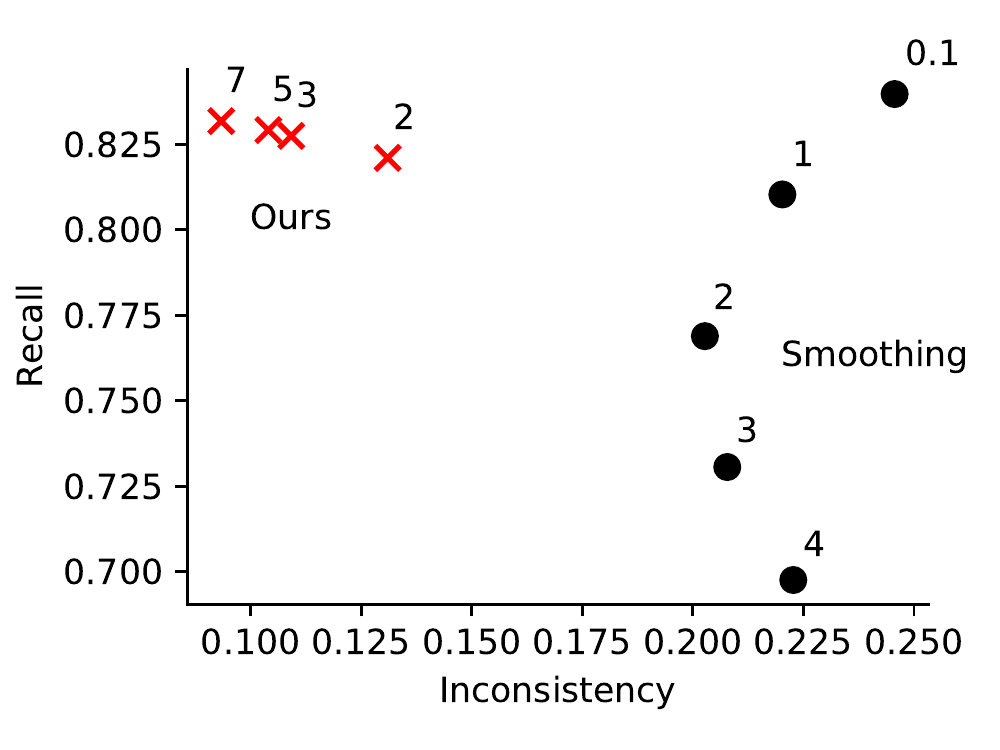}

\caption{\textbf{Our model vs Temporal smoothing of the baseline model} We compare recall (higher is better) and inconsistency (lower is better) for our multi-input model with various number of input frames (Red), against the results of temporally smoothing the baseline 'naive' frame-wise segmentation model in post-process with various gaussian kernel sigmas (Black).} 
\label{fig:smoothing_baseline}
\end{figure}

Another common approach to improving temporal stability is to take the output of a model trained on still images and apply a Gaussian filter in the time dimension. This form of temporal smoothing can indeed improve the consistency of video segmentation, and is computationally cheap. However, improving consistency should not be pursued without preserving accuracy, as one can always return the trivial solution of a constant mask. One can ask then whether our method provides a solution that is outside the consistency-accuracy Pareto front of temporal smoothing. As shown in \Cref{fig:smoothing_baseline}, smoothing doesn't necessarily improve consistency, and monotonously degrades accuracy. In contrast, our method shows that one can improve temporal consistency without losing accuracy.

\subsection{Postprocessing with 3D Guided Filter}
\label{sec:guided_filter}

To improve consistency further we experimented with post processing using a guided filter \cite{he2012guided}, extended to 3D (2D+time), with the expectation that such a local filter will further improve the agreement between model predictions on consecutive frames. In particular, we implemented the weighted guided filter of~\cite{liba2020sky}, which assigns a smaller weight to low confidence pixels. We extend it to 3D in a straightforward manner, by stacking the output masks as a volume and using the stack of frames as the guide, a process that takes a few seconds for a short clip. For multi-class predictions, we filter each mask independently, after softmax but before binarization, and set the label of a pixel to the class with highest value after filtering.

The effect of applying the 3D weighted guided filter (WGF) is summarized in \Cref{tab:guidedfilt}. This additional post processing step improves consistency significantly. It also shows small improvements in smoothness, although guided filters usually tend to sharpen, so we attribute this improvement to the masks being more consistent. This improved consistency comes with a price of degraded recall. When we apply the WGF with temporal kernel size equal to 1, which is effectively a 2D WGF, it has negligible effect on consistency and smoothness (but slightly better recall). This demonstrates the effectiveness of applying the filter on the temporal dimension as well.

\begin{table}
  \centering

\begin{tabular}{lcccc}
    \toprule
     & Recall$\uparrow$ & Smoothness $\downarrow$ &  Inconsistency $\downarrow$  \\
    \midrule
    Ours  & 0.831 & 3.91 & 0.093  \\
    + 2d WGF & \textbf{0.833} & 4.11 & 0.093 \\
    + 3d WGF  & 0.826 & \textbf{3.17} & \textbf{0.075} \\
    \bottomrule
\end{tabular}

  \caption{
      \textbf{Effect of weighted guided filter (WGF) on temporal consistency.}
      Computed with our method with 7 inputs over the DAVIS dataset.
      3D WGF improves consistency with some cost on recall. We used spatial kernel of size 8 both in 2D and 3D WGF, and a temporal kernel of size 4 for the 3D WGF.
      }
  \label{tab:guidedfilt}
\end{table}

\subsection{Temporal Consistency for Video Models}
\begin{figure}
\centering
    \includegraphics[width=0.65\linewidth]{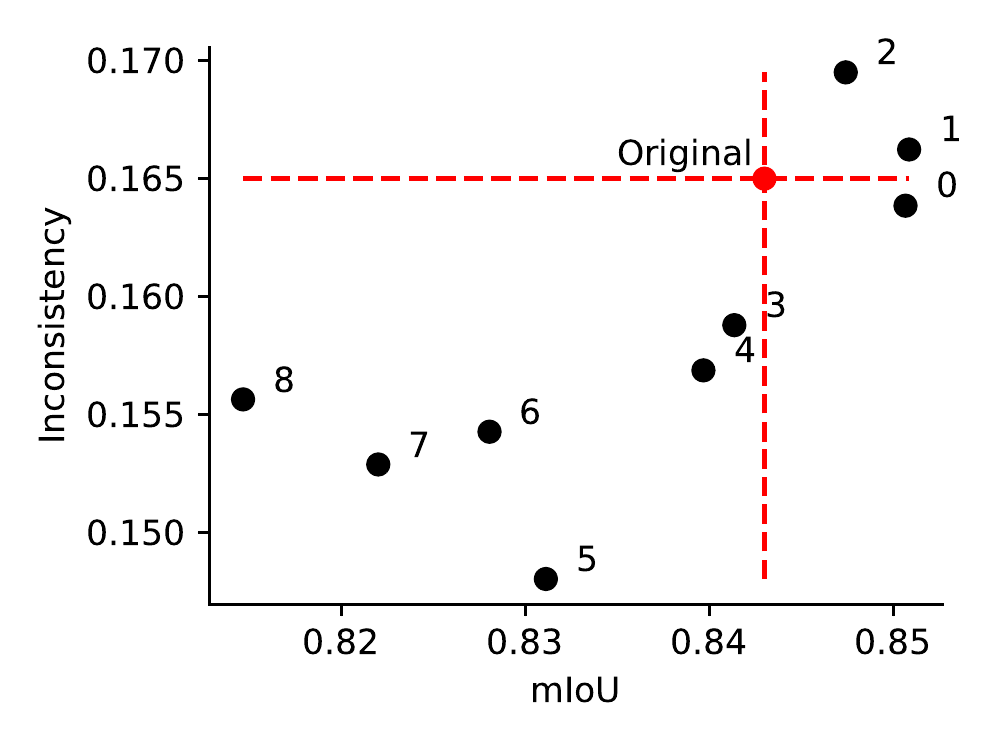}

  \caption{\textbf{Improving consistency of video segmentation models.} Original (red point) corresponds to the model presented by Mahadevan et al. \cite{mahadevan2020making}. The black points indicate different training epochs with our consistency loss.}
  \label{fig:video_seg}
\end{figure}

Having shown in detail how an image segmentation model can be turned into a temporally stable video model, we briefly turn to the question of temporal stability in models that were trained on video data. We tested the straight-forward approach of fine-tuning a video model on the same data it was originally trained on (without additional data), but adding our proposed consistency loss. Starting from a state-of-the-art video segmentation model \cite{mahadevan2020making} and training as described above, one can get a variety of possible solutions. In particular, \Cref{fig:video_seg} shows that one can explore the accuracy-consistency trade-off by changing the number of training epochs. We leave for future work the attempt to improve video segmentation models without access to the original labeled video data by using a teacher student approach with our consistency loss (as we showed in \Cref{sec:decoder_training}).

\section{Conclusions}

In this paper we tackled the task of training temporally stable video segmentation models. Given the difficulty and cost of collecting and annotating video datasets, we proposed a training method that uses still image datasets rather than annotated videos. We based our approach on a consistency measure for video segmentation that we found to be well aligned to the human perception of temporal stability. We used it, combined with image dataset refinement and 3D guided filter post-processing, to fine-tune existing segmentation models as well as to train a multi-input network from scratch, improving stability in all cases while maintaining satisfying recall rates.
While this approach is quite computationally expensive and can not be used for online inference, it can be used to produce on demand temporally consistent labeled videos which can be used for training a student video segmentation network.


\section{Acknowledgments}
We thank Tal Arkushin for the helpful discussion. Support by the ISF and the Gatsby Foundation is gratefully acknowledged.

{\small 
\bibliographystyle{ieee_fullname}
\bibliography{egbib}
}

\end{document}